\documentclass[letterpaper]{article} % DO NOT CHANGE THIS
\usepackage{aaai24}  % DO NOT CHANGE THIS
\usepackage{times}  % DO NOT CHANGE THIS
\usepackage{helvet}  % DO NOT CHANGE THIS
\usepackage{courier}  % DO NOT CHANGE THIS
\usepackage[hyphens]{url}  % DO NOT CHANGE THIS
\usepackage{graphicx} % DO NOT CHANGE THIS
\usepackage{natbib}  % DO NOT CHANGE THIS AND DO NOT ADD ANY OPTIONS TO IT
\usepackage{caption} % DO NOT CHANGE THIS AND DO NOT ADD ANY OPTIONS TO IT
\usepackage{algorithm}
\usepackage{algorithmic}
\usepackage{amsfonts}
\usepackage{booktabs}
\usepackage{multirow}
\usepackage{makecell}
\usepackage{color}

\usepackage{newfloat}
\usepackage{listings}
\DeclareCaptionStyle{ruled}{labelfont=normalfont,labelsep=colon,strut=off} % DO NOT CHANGE THIS
\floatstyle{ruled}
\newfloat{listing}{tb}{lst}{}
\floatname{listing}{Listing}
\title{Deep Contrastive Graph Learning with Clustering-Oriented Guidance}
\author{
	Mulin Chen\textsuperscript{\rm 1,\rm 2}, Bocheng Wang\textsuperscript{\rm 1,\rm 2}, Xuelong Li\textsuperscript{\rm 1,\rm 2}\thanks{Corresponding author.}
}
\affiliations{
    \textsuperscript{\rm 1} School of Artificial Intelligence, OPtics and
    ElectroNics (iOPEN),  \\
    Northwestern Polytechnical University, Xi'an 710072, Shanxi, China \\
    \textsuperscript{\rm 2} Key Laboratory of Intelligent Interaction and Applications, Ministry of Industry and Information Technology, \\
    Northwestern Polytechnical University, Xi'an 710072, Shanxi, China \\
    chenmulin@nwpu.edu.cn, wangbocheng@mail.nwpu.edu.cn, li@nwpu.edu.cn
    
}

\usepackage{bibentry}
\begin{document}

\maketitle

\begin{abstract}
	Graph Convolutional Network (GCN) has exhibited remarkable potential in improving graph-based clustering. To handle the general clustering scenario without a prior graph, these models estimate an initial graph beforehand to apply GCN. Throughout the literature, we have witnessed that 1) most models focus on the initial graph while neglecting the original features. Therefore, the discriminability of the learned representation may be corrupted by a low-quality initial graph; 2) the training procedure lacks effective clustering guidance, which may lead to the incorporation of clustering-irrelevant information into the learned graph. To tackle these problems, the Deep Contrastive Graph Learning (DCGL) model is proposed for general data clustering. Specifically, we establish a pseudo-siamese network, which incorporates auto-encoder with GCN to emphasize both the graph structure and the original features. On this basis, feature-level contrastive learning is introduced to enhance the discriminative capacity, and the relationship between samples and centroids is employed as the clustering-oriented guidance. Afterward, a two-branch graph learning mechanism is designed to extract the local and global structural relationships, which are further embedded into a unified graph under the cluster-level contrastive guidance. Experimental results on several benchmark datasets demonstrate the superiority of DCGL against state-of-the-art algorithms.
\end{abstract}

\section{Introduction}

Graph-based clustering achieves dominant performance due to its capability on capturing data manifold. It uses an undirected graph to learn the pair-wise relations between samples, and infers the cluster labels from the adjacency structure.  Recently, neural networks are widely used in data clustering \cite{DEC,SpectralNet,SDCN,deepClustering}. Among a variety of network structures, Graph Convolutional Network (GCN) \cite{GCN} has shown a prominent ability on exploring the relationship among graph-type data, and realized good results on graph-based clustering \cite{AnchorGAE}. 

GCN-based clustering methods learn the data representation with the input graph and data matrix. For general clustering tasks, there is no prior graph that records the relationships of samples. Hence, an initial graph needs to be constructed as the input of GCN. Although this approach has demonstrated promising results, it gives rise to some issues. Firstly, as observed in the Graph Auto-Encoder (GAE) framework \cite{AdaGAE,AGCNN} and a series of graph-structured data clustering models \cite{DAEGC,DFCN,DCRN}, the training procedure relies on the input graph heavily, and pays insufficient attention to the original features. Therefore, if there are many connected samples from different classes in the initial graph, the embedding obtained by multilayer aggregation would be ambiguous, leading to a loss of discriminative capacity. This phenomenon is called as representation collapse. Secondly, GCN-based models emphasize the representational ability of the learned graph embedding, but lack clustering-oriented guidance, such as the distribution of centroids. As a result, the output of GCN may contain clustering-irrelevant information, while the cluster structure is not fully captured, which is essential for data clustering. 

\begin{figure*}[t]
	\centering
	\includegraphics[width=0.95\textwidth]{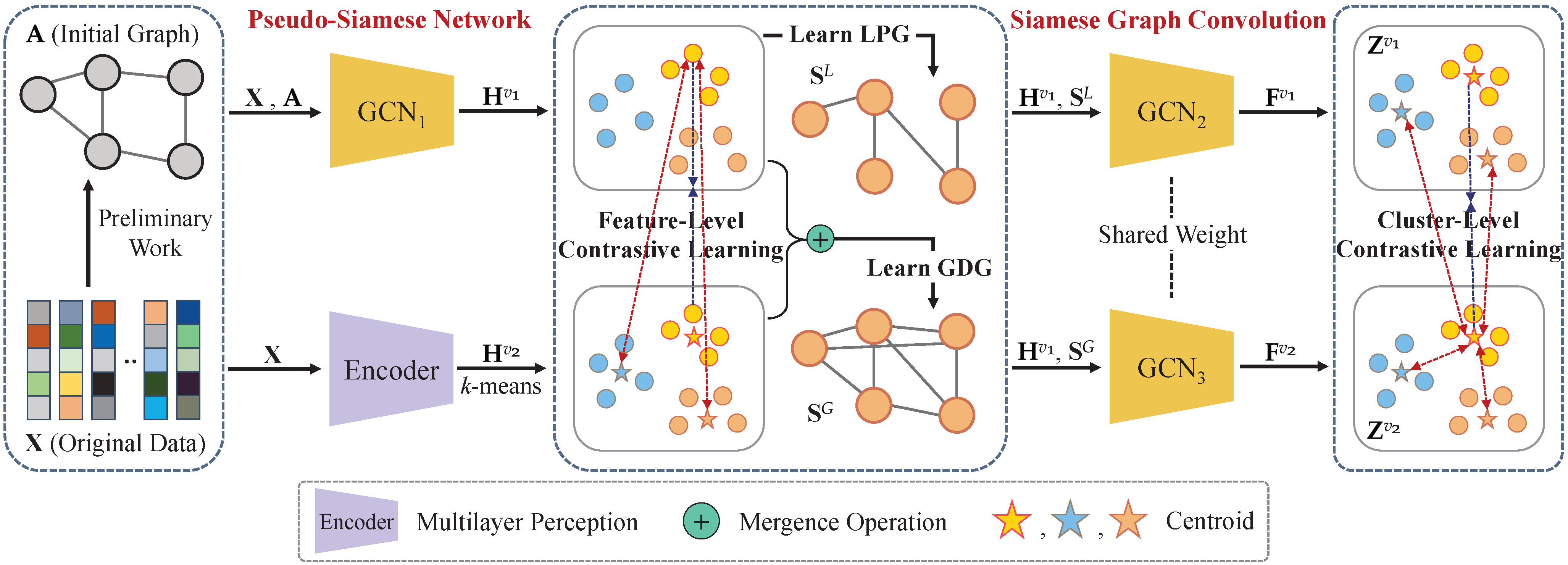} 
	\caption{Overview of DCGL. Note that the decoder for $\mathbf{H}^{v_2}$ is omitted in this pipeline. First, we construct an initial graph to enable GCN. Second, we generate two latent representations by the pseudo-siamese network. Third, we derive two adjacency graphs from different perspectives, and then compute their cluster-level graph embeddings. Afterward, the framework is updated with the joint loss shown in Eq. (\ref{eq:overall_loss}). The final result is obtained by performing spectral clustering on the converged $\mathbf{S}^L$.}
	\label{fig:pipline}
\end{figure*}

To address these issues, we propose the Deep Contrastive Graph Learning (DCGL) model for the general clustering task without a prior graph. As shown in Fig. \ref{fig:pipline}, DCGL consists of a pseudo-siamese network, wherein one branch handles both the data matrix and the input graph with GCN, while the other branch encodes the data matrix with an encoder. During this stage, feature-level contrastive learning models the distances between the estimated centroids and samples, such that the discriminative characteristics are explored. Then, the outputs are employed to construct the Local Propinquity Graph (LPG) and Global Diffusion Graph (GDG) with different neighborhood sizes respectively. After that, the graphs are mapped into the cluster space, where cluster-level contrastive learning is used to push away the different centroids. In this way, the cluster structure is better maintained. The main contributions of this paper are summarized as follows:

\begin{itemize}
	\item A deep unsupervised graph learning model is established to cluster general data, which takes a pseudo-siamese network to analyze structural dependencies and node attributes concurrently. By considering the original features explicitly and updating the learned graph adaptively, the reliance on the initial graph is alleviated.
	
	\item Two clustering-oriented contrastive learning strategies are devised to provide training guidance. Feature-level contrastive learning aims to preserve the discriminability of each node, and cluster-level contrastive learning promotes a clear distribution of centroids.
	
	\item The graph learning procedure is divided into two branches to capture the data manifold from local and global perspectives respectively. Both the local and global graphs are mapped into the cluster space to search a unified cluster structure.
	
\end{itemize}

\section{Related Work}

\subsection{Graph Structure Learning}

Graph learning is an important topic in many practical applications \cite{SCGC, duomotai, anFang}. $K$-Nearest Neighbor (KNN) is the most traditional graph learning way. This strategy builds the adjacency relation based on the sample distance, and then calculates the weight by a specific formula \cite{GNMF, LADA, MPF}. The graph from simple KNN reserves limited manifold information, so some advanced graph learning theories are raised, such as adaptive neighbor learning \cite{CAN,RGC}, self-expression \cite{SSC,SCSK}, and consistency propagation \cite{huang2022multi,consistencypropagation}.  

In recent years, Graph Neural Network (GNN) has presented reliable competence in deep graph structure learning. GLCN \cite{GLCN} exploits a unified framework for executing graph learning and graph convolution. IDGL \cite{IDGL} leverages graph regularization terms to enhance learning quality. Pro-GNN \cite{Pro-GNN} employs alternating optimization to realize the mutual regularization of graph learning and network update. GRCN \cite{GRCN} predicts edges and regulates weights adaptively by residual link. VIB-GSL \cite{VIB-GSL} quotes a novel theory of variational information bottleneck to structure an adjacency graph. The above influential models have shown impressive accuracy in experiments. However, most of the deep models are tailored for some kind of supervised scenario, which is usually node classification. SUBLIME \cite{SUBLIME} is described as the first attempt at unsupervised deep graph learning. In this paper, we plan to develop an unsupervised procedure to support graph-based clustering.

\subsection{GNN-Based Data Clustering}

To date, a series of GNN-based models have achieved excellent performance in graph-structured data clustering \cite{GraphSurvey}. DAEGC \cite{DAEGC} uses a new GAE based on graph attention network to learn pivotal node features. SDCN \cite{SDCN} performs GCN and auto-encoder simultaneously for better cluster learning. DFCN \cite{DFCN} applies a representation fusion module to integrate the topology structure and node attributes. DCRN \cite{DCRN} utilizes the graph augmentation technology to explore different levels of latent characteristics, and then merges them into a consensus cluster representation. 

In the above methods, the final cluster embedding relies on the input graph strongly, which means that the initial graph has a decisive influence on clustering performance. Therefore, the aforesaid models are not appropriate for the general clustering scenario without a reliable graph structure. Very lately, some modified GAEs \cite{AdaGAE,AnchorGAE} are established to handle general data, where the adjacency relation is updated automatically. However, GAE fails to preserve the discriminative characteristics within the original features. Moreover, most models lack clustering-oriented training guidance, and the learned graph may be unsuitable for clustering.

\subsection{Contrastive Learning}

As an effective tool in visual processing \cite{contrastiveVis}, contrastive learning has been applied to many tasks in the branch of GNN successfully, such as graph representation learning \cite{CMVRL,GCA,SUGRL} and graph-structured data clustering \cite{contrastGC1,contrastGC2,CCGC}. In practical application, contrastive learning is usually joined with graph augmentation. Among them, graph augmentation aims to derive multiple data views from node-level and topology-level. Contrastive learning is used to maximize the mutual information between different views to generate the final node representation, which means pulling correlative semantic contents closer while pushing outlying features away. In this article, we introduce clustering-oriented guidance into contrastive learning methodologies, ensuring enhanced discriminability between samples and pursuing a distinguishable cluster structure.

\section{Deep Contrastive Graph Learning}
In this section, the proposed model is introduced. The pipeline of DCGL is illustrated in Fig. \ref{fig:pipline}. It uses a presudo-siamese network to construct the local and global graphs, and maps them into the cluster space by siamese graph convolution.

\textbf{Notations:} in this paper, vectors and matrices are written in lowercase and uppercase letters respectively. For a data matrix $\mathbf{X}$, both $\mathbf{x}_{i}$ and $\mathbf{X}_{i}$ represent the $i$-th sample. $||\mathbf{x}_{i}||_{1}$, $||\mathbf{x}_{i}||_{2}$, and $||\mathbf{X}||_{\mathrm{F}}$ denote $\ell_{1}$, $\ell_{2}$, and Frobenius norm respectively.

\subsection{Preliminary Work}

Given a dataset $\mathbf{X} = [\mathbf{x}_1, \mathbf{x}_2, \mathbf{x}_3,\cdots, \mathbf{x}_n]\in\mathbb{R}^{n\times m}$ with $c$ categories, it is necessary to construct an initial graph as the input of GCN.

The initial graph indicates the relationship between two samples by the corresponding weight. Intuitively, a small Euclidean distance reflects a large weight. Therefore, the objective is expressed as
\begin{equation}
	\label{eq:gl}
	\begin{array}{c}
		\mathop {\min }\limits_\mathbf{A} {\rm{ }}\mathop \sum \limits_{i,j}^n {\rm{||}}{\mathbf{x}_i} - {\mathbf{x}_j}{\rm{||}}_2^2{a_{ij}} + \gamma {\rm{||}}\mathbf{A}{\rm{||}}_{\rm{F}}^2,\\
		s.t.{\rm{ }}{\rm{ ||}}{\mathbf{a}_i}{\rm{|}}{{\rm{|}}_1} = 1,{\rm{ }}0 \le {\mathbf{a}_i} \le 1,
	\end{array}
\end{equation}
where $\gamma$ is the trade-off parameter, $\mathbf{A} \in \mathbb{R}^{n\times n}$ is the desired graph, and $\mathbf{a}_i$ refers to the $i$-th column of $\mathbf{A}$. The second term aims to avoid the trivial solution $\mathbf{A}=\mathbf{I}$.

Problem (\ref{eq:gl}) can be decoupled into each column of $\mathbf{A}$ separately. Denoting a column vector $\mathbf{d}_{i} \in \mathbb{R}^{n\times 1}$ that satisfies $d_{ij}=||\mathbf{x}_{i}-\mathbf{x}_{j}||_{2}^{2}$, the sub-problem for $\mathbf{a}_{i}$ can be written as
\begin{equation}
	\label{eq:gl_sub}
	\begin{array}{c}
		\mathop {\min {\rm{ }}}\limits_{{\rm{||}}{\mathbf{a}_i}{\rm{|}}{{\rm{|}}_1} = 1,{\rm{ }}0 \le {\mathbf{a}_i} \le 1{\rm{ }}} \left\| {{\mathbf{a}_i} + \frac{1}{{2\gamma }}\mathbf{d}_{i}} \right\|_2^2.
	\end{array}
\end{equation}

Define $\mathbf{w}_{i}$ as the result of sorting $\mathbf{d}_{i}$ in ascending order. Stipulating that only the weights of $k$ neighbors are updated, problem (\ref{eq:gl_sub}) has a closed-form solution as
\begin{equation}
	\label{eq:gl_solution}
	\begin{array}{c}
		{a_{ij}} = \frac{{{{({w_{i(k + 1)}} - {w_{ij}})}_ + }}}{{k{w_{i(k + 1)}} - \sum\nolimits_u^k {{w_{iu}}} }},
	\end{array}
\end{equation}
where $(\cdot)_{+}$ indicates $max(\cdot,0)$. The final graph is obtained by $(\mathbf{A}+\mathbf{A}^{\mathrm{T}})/2$, and the mathematical deduction can be found in \cite{CAN}. 

With the scheme of Eq. (\ref{eq:gl}), each sample connects with its neighbors adaptively. Moreover, the specific closed-form solution avoids the parameter selection of $\gamma$, and only the neighbor number $k$ needs to be tuned.

\subsection{Pseudo-Siamese Network}

After building the previous stage, the input for GCN is obtained, including data representation $\mathbf{X}$ and initial graph $\mathbf{A}$. GCN mainly pays attention to the data relationship, but neglects the samples' original attributes, and encounters the risk of representation collapse. To preserve the original features, we parallel GCN with an auto-encoder, leading to a pseudo-siamese network, which is formulated as
\begin{equation}
	\label{eq:pseudo_siamese}
	\begin{array}{c}
		{\mathbf{H}^{{v_1}}} = \mathrm{GCN}_{1}(\mathbf{X},\widehat{\mathbf{A}}),\\
		{\mathbf{H}^{{v_2}}} = \mathrm{Encoder}(\mathbf{X}),\widehat{\mathbf{X}} = \mathrm{Decoder}(\mathbf{H}^{{v_2}}),
	\end{array}
\end{equation}
where $\widehat{\mathbf{A}}$ is the normalized graph, $\widehat{\mathbf{A}}=\widetilde{\mathbf{D}}^{-\frac{1}{2}}\mathbf{\widetilde{\mathbf{A}}}\widetilde{\mathbf{D}}^{-\frac{1}{2}}$, $\widetilde{\mathbf{A}}=\mathbf{A}+\mathbf{I}$, $\widetilde{\mathbf{D}} \in \mathbb{R}^{n\times n}$ refers to the diagonal degree matrix that satisfies $\widetilde{\mathbf{D}}_{i i}=\sum_{j}^{n} \tilde{\mathbf{A}}_{i j}$, and $\mathbf{H}^{v_1} \in \mathbb{R}^{n\times l}$ and $\mathbf{H}^{v_2} \in \mathbb{R}^{n\times l}$ captures the structure and attributes respectively.
The classical Mean-Square Error is used to train the auto-encoder
\begin{equation}
	\label{eq:ae_loss}
	\begin{array}{c}
		{{\cal L}_{AE}} = \frac{1}{2n}\sum\limits_i^n {{{\left\| {{\mathbf{x}_{i}} - {{\widehat{\mathbf{x}}}_i}} \right\|}_2^2}}.
	\end{array}
\end{equation}

The above formulation focuses on data representation, and the training procedure lacks cluster-oriented guidance. Therefore, we design the following feature-level contrastive learning to improve the representations' adaptability to clustering tasks.

\subsubsection{Feature-Level Contrastive Learning.} In contrastive learning, the selection of positive and negative samples determines the final effect directly. For an anchor $\mathbf{H}_{i}^{v_1}$, the universal consensus \cite{GCA,SUBLIME} is to treat its embedding in the other branch as the positive sample ($\mathbf{H}_{i}^{v_2}$), and all the rest as negative samples, which may misjudges too many samples in the same category as negative. Differently, we employ a rough clustering result to guide the construction of sample pairs. For each anchor, the centroids of other clusters are taken as negative ones.

Specifically, $k$-means is executed on $\mathbf{H}_{i}^{v_2}$ to acquire $c$ centroids. Given any node $\mathbf{H}_{i}^{v_1}$, the centroids of other clusters constitute the negative sample set $\mathbf{P}^{(i)} \in \mathbb{R}^{(c-1)\times l}$. On this basis, the InfoNCE loss \cite{InfoNCE} is employed for feature-level contrastive learning
\begin{equation}
	\label{eq:loss_flcl}
	\begin{array}{c}
		{{\cal L}_{FL}} = \frac{1}{n}\sum\limits_i^n {\left( { - \log \frac{{{e^{\theta \left( {\mathbf{H}_i^{{v_1}},\mathbf{H}_i^{{v_2}}} \right)/\tau }}}}{{{e^{\theta \left( {\mathbf{H}_i^{{v_1}},\mathbf{H}_i^{{v_2}}} \right)/\tau }} + \sum\limits_j^{c - 1} {{e^{\theta \left( {\mathbf{H}_i^{{v_1}},\mathbf{P}^{(i)}_{j}} \right)/\tau }}} }}} \right)} ,
	\end{array}
\end{equation}
where $\theta(\cdot,\cdot)$ computes the cosine similarity, and $\tau$ is the temperature parameter. 

Eq.~(\ref{eq:loss_flcl}) pushes samples away from incorrect centroids, ensuring the discriminability of the learned representation.

\subsubsection{Local Propinquity Graph Learning.} Replacing the $\mathbf{x}_i$ in Eq. (\ref{eq:gl}) with the learned $\mathbf{H}_i^{{v_1}}$, a new graph $\mathbf{S}^L$ is derived. This graph learning problem can be converted to trace form as
\begin{equation}
	\label{eq:loss_gl}
	\begin{array}{c}
		{{\cal L}_{GL}} = \mathrm{Tr}\left( {{{({\mathbf{H}^{{v_1}}})}^\mathrm{T}}{\mathbf{L_S}}{\mathbf{H}^{{v_1}}}} \right) + \frac{\gamma}{2} \mathrm{Tr}\left(\mathbf{S}^L(\mathbf{S}^L)^\mathrm{T}\right),
	\end{array}
\end{equation}
where $\mathrm{Tr}(\cdot)$ denotes the trace of a square matrix, and $\mathbf{L_S} \in \mathbb{R}^{n\times n}$ is the Laplace matrix of $\mathbf{S}^L$. The above formulation infers the manifold structure based on the pair-wise similarity, so we term $\mathbf{S}^L$ as Local Propinquity Graph (LPG). With a small value of $k$, only the very close samples will be connected, such that the final graph contains fewer incorrect connections. To detect more potential connections, we adopt a staged growth for the neighbor number, which means adding $k$ with a fixed increment after a constant interval.

\subsection{Siamese Graph Convolution}

The optimization of LPG focuses on the local dependencies between nodes. Considering that some samples of the same class may are not neighbors in Euclidean space, we propose to learn the global similarity with an extra graph, and fuse both the local and global graphs to achieve an agreement on cluster structure.

\subsubsection{Global Diffusion Graph Learning.} Contrary to LPG, the new graph encodes the global topological relationship of samples, which requires more links between nodes. Firstly, the structural and attributed nodes are combined to generate a comprehensive representation
\begin{equation}
	\label{eq:mergence}
	\begin{array}{c}
		\mathbf{H} = \frac{1}{2}({\mathbf{H}^{{v_1}}} + {\mathbf{H}^{{v_2}}}).
	\end{array}
\end{equation}

Then, the integrated $\mathbf{H}$ is used as the data matrix to construct a graph $\mathbf{G}$ with Eq. (1), which further exploits the original features. The neighbor number for $\mathbf{G}$ is fixed as the upper bound $\left\lfloor {\frac{n}{c}} \right\rfloor$ of the learning procedure of LGP, where $\lfloor {\cdot} \rfloor$ denotes the operation of round down.

Finally, Personalized PageRank graph diffusion \cite{CMVRL} is introduced to transport the topological similarity through nodes, so as to further capture the global structure. The mathematical expression is
\begin{equation}
	\label{eq:gdg_diffusion}
	\begin{array}{c}
		\mathbf{S}^{G} = \sum\limits_{i=0}^\infty  {\lambda {{\left( {1 - \lambda } 	\right)}^i}{\left({\mathbf{D_G}}^{ - \frac{1}{2}}\mathbf{G}{\mathbf{D_G}}^{ - \frac{1}{2}}\right)^i}},
	\end{array}
\end{equation}
where $\lambda$ is the transport rate, and $\mathbf{D_G} \in \mathbb{R}^{n\times n}$ is the degree matrix of $\mathbf{G}$. The corresponding closed-form solution is
\begin{equation}
	\label{eq:gdg_ppr}
	\begin{array}{c}
		{\mathbf{S}^{G}} = \lambda {\left[ {{\mathbf{I_n}} - (1 - \lambda )\mathbf{D_G}^{ - \frac{1}{2}}\mathbf{G}\mathbf{D_G}^{ - \frac{1}{2}}} \right]^{ - 1}},
	\end{array}
\end{equation}
where $\mathbf{I_n} \in \mathbb{R}^{n\times n}$ is the identity matrix.

$\mathbf{S}^{G}$ is called Global Diffusion Graph (GDG). Compared with the straightforward inner product, the construction of GDG focalizes manifold information of each cluster, rather than the holistic sample space.

\subsubsection{Cluster-Level Contrastive Learning.} With both the local graph $\mathbf{S}^{L}$ and global graph $\mathbf{S}^{G}$, contrastive learning is performed within the cluster space.

We assume that both graphs should share the same cluster structure. To this end, the graphs and embeddings are fed into a siamese GCN to get the cluster indicators
\begin{equation}
	\label{eq:gcn_ge}
	\begin{array}{c}
		{\mathbf{F}^{v_1}} = \mathrm{GCN}_{2}({\mathbf{H}^{v_1}},\widehat{\mathbf{S}}^{L}), \\
		{\mathbf{F}^{v_2}} = \mathrm{GCN}_{3}({\mathbf{H}^{v_1}},{\widehat{\mathbf{S}}^{G}}),
	\end{array}
\end{equation}
where $\widehat{\mathbf{S}}^L$ and $\widehat{\mathbf{S}}^{G}$ are the normalized graphs of $\mathbf{S}^L$ and $\mathbf{S}^{G}$. The softmax function is employed to activate the last layer of GCN. In this way, each element of $\mathbf{F}^{v_1} \in \mathbb{R}^{n\times c}$ and $\mathbf{F}^{v_2} \in \mathbb{R}^{n\times c}$ records the probability that a sample belongs to a cluster. To emphasize the optimization of LPG, both $\mathrm{GCN}_2$ and $\mathrm{GCN}_3$ use the structural representation $\mathbf{H}^{v_1}$ as input.

Then, the cluster-level embedding is calculated as
\begin{equation}
	\label{eq:lc}
	\begin{array}{c}
		{\mathbf{Z}^{v_1}} = {({\mathbf{F}^{v_1}})^\mathrm{T}}{\mathbf{H}^{v_1}},{\mathbf{Z}^{v_2}} = {({\mathbf{F}^{v_2}})^\mathrm{T}}{\mathbf{H}^{v_1}},
	\end{array}
\end{equation}
where $\mathbf{Z}^{v_1} \in \mathbb{R}^{c\times l}$ and $\mathbf{Z}^{v_2} \in \mathbb{R}^{c\times l}$ consist of the cluster centroids. Eq. (12) projects the cluster indicators into centroid matrices. Afterward, the centroids are used as anchors, and contrastive learning is conducted to search a shared centroid distribution
\begin{equation}
	\label{eq:cl_loss}
	\begin{array}{c}
		{{\cal L}_{CL}} = \frac{1}{{2c}}\sum\limits_i^c {\left[ {\Omega (\mathbf{Z}^{v_1}_{i},\mathbf{Z}^{v_2}_{i}) + \Omega (\mathbf{Z}^{v_2}_{i},\mathbf{Z}^{v_1}_{i})} \right]},
	\end{array}
\end{equation}
where
\begin{equation}
	\label{eq:cl_single}
	\begin{array}{c}
		\Omega (\mathbf{Z}^{v_1}_{i},\mathbf{Z}^{v_2}_{i}) =  - \log \frac{{{e^{\theta \left( {\mathbf{Z}^{v_1}_{i},\mathbf{Z}^{v_2}_{i}} \right)/\tau }}}}{{\sum\limits_j^c {{e^{\theta \left( {\mathbf{Z}^{v_1}_{i},\mathbf{Z}^{v_2}_{j}} \right)/\tau }}} }} + \sum\limits_{j \ne i}^c {{e^{\theta \left( {\mathbf{Z}^{v_1}_{i},\mathbf{Z}^{v_1}_{j}} \right)/\tau }}}.
	\end{array}
\end{equation}

Eq.~(\ref{eq:cl_loss}) pushes the centroids of different clusters away, and ensures an explicit cluster structure. The siamese graph convolution module does not involve the initial graph $\mathbf{A}$, so the impact of the initial graph quality is  reduced. Moreover, both cluster-level and feature-level contrastive learning employ centroids to leverage cluster structure, and further provide training guidance to facilitate clustering.

\begin{algorithm}[tb]
	\caption{\textbf{DCGL}}
	\label{alg:dcgl}
	\textbf{Input}: Data matrix $\mathbf{X}$ with $n$ samples, cluster number $c$, initial neighbor number $k$, neighbor number update interval $t$, maximum iterations $iter$, parameters $\alpha$, $\beta$, $\gamma$.\\
	\textbf{Output}: Clustering result for $\mathbf{X}$.
	\begin{algorithmic}[1] 
		\STATE Let $k^{init} = k, i = 1$.
		\STATE Compute $\mathbf{A}$ by solving problem (\ref{eq:gl}) with $\mathbf{X}$ and $k^{init}$.
		\WHILE{$k \leq \lfloor {\frac{n}{c}} \rfloor$ \textbf{and} $i \leq iter$}
		\STATE Obtain $\mathbf{H}^{v_1}$, $\mathbf{H}^{v_2}$, and $\widehat{\mathbf{X}}$ by pseudo-siamese network.
		\STATE Perform $k$-means on $\mathbf{H}^{v_2}$ to obtain $c$ centroids.
		\STATE Compute $\mathbf{S}^L$ by solving problem (\ref{eq:gl}) with $\mathbf{H}^{v_1}$ and $k$.
		\STATE Compute $\mathbf{H}$ with Eq. (\ref{eq:mergence}).
		\STATE Compute $\mathbf{G}$ by solving problem (\ref{eq:gl}) with $\mathbf{H}$ and $\lfloor {\frac{n}{c}} \rfloor$.
		\STATE Compute $\mathbf{S}^{G}$ with Eq. (\ref{eq:gdg_ppr}).
		\STATE Obtain $\mathbf{F}^{v_1}$ and $\mathbf{F}^{v_2}$ by siamese GCN.
		\STATE Compute $\mathbf{Z}^{v_1}$ and $\mathbf{Z}^{v_2}$ with Eq. (\ref{eq:lc}).
		\STATE Compute ${\cal L}_{AE}$, ${\cal L}_{FL}$, ${\cal L}_{GL}$, and ${\cal L}_{CL}$, respectively.
		\STATE Update model by minimizing Eq. (\ref{eq:overall_loss}) with Adam.
		\IF {$i\%t==0$}
		\STATE Let $k = k+k^{init}$.
		\ENDIF
		\STATE Let $i = i + 1$.
		\ENDWHILE
		\STATE Perform spectral clustering on $\mathbf{S}^L$.
		\RETURN Cluster labels.
	\end{algorithmic}
\end{algorithm}

\begin{figure*}[t]
	\center
	\begin{minipage}[t]{0.14\textwidth}
		\center
		\includegraphics[width=1\textwidth]{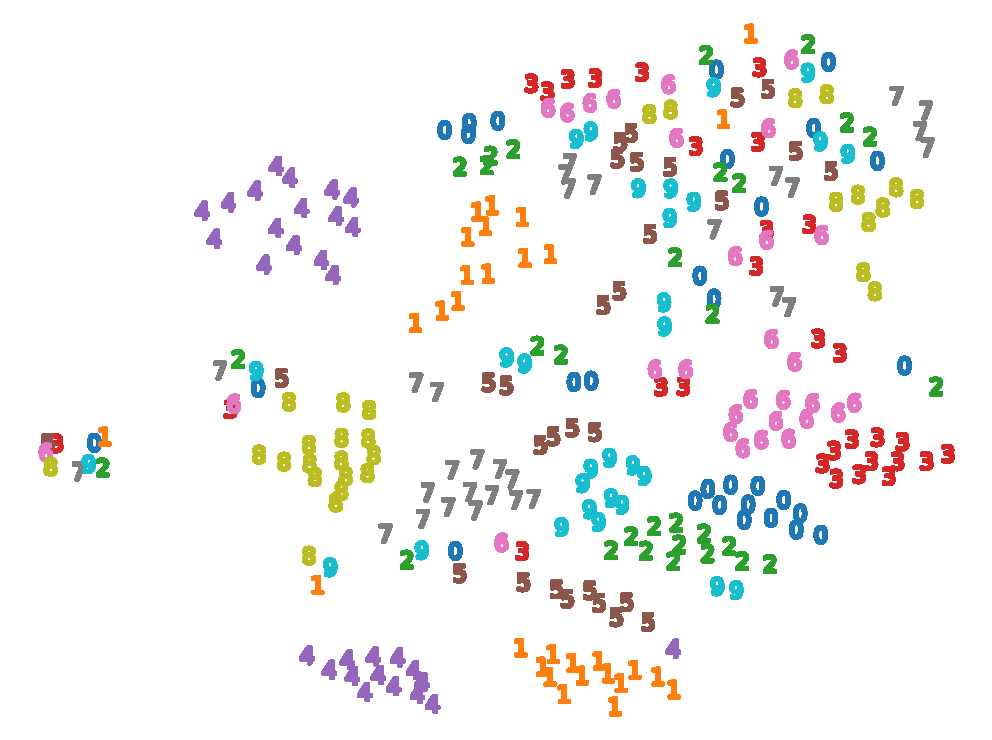} 
		\centerline{Raw}
	\end{minipage}
	\hspace{10pt}
	\begin{minipage}[t]{0.14\textwidth}
		\center
		\includegraphics[width=1\textwidth]{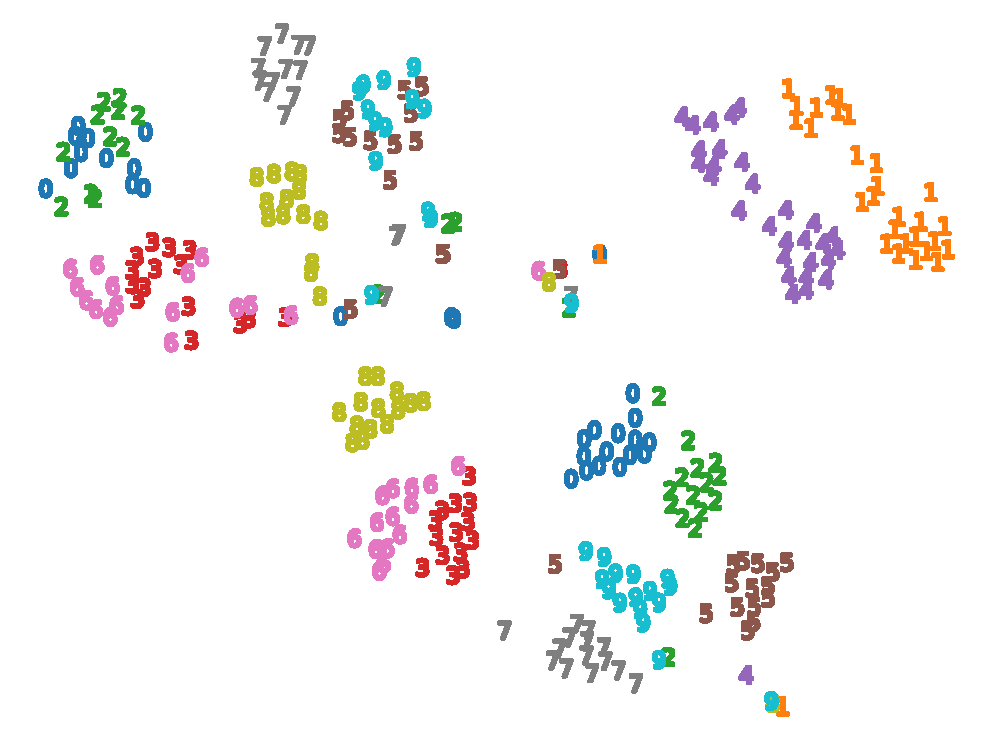} 
		\centerline{Epoch-6}
	\end{minipage}
	\hspace{10pt}
	\begin{minipage}[t]{0.14\textwidth}
		\center
		\includegraphics[width=1\textwidth]{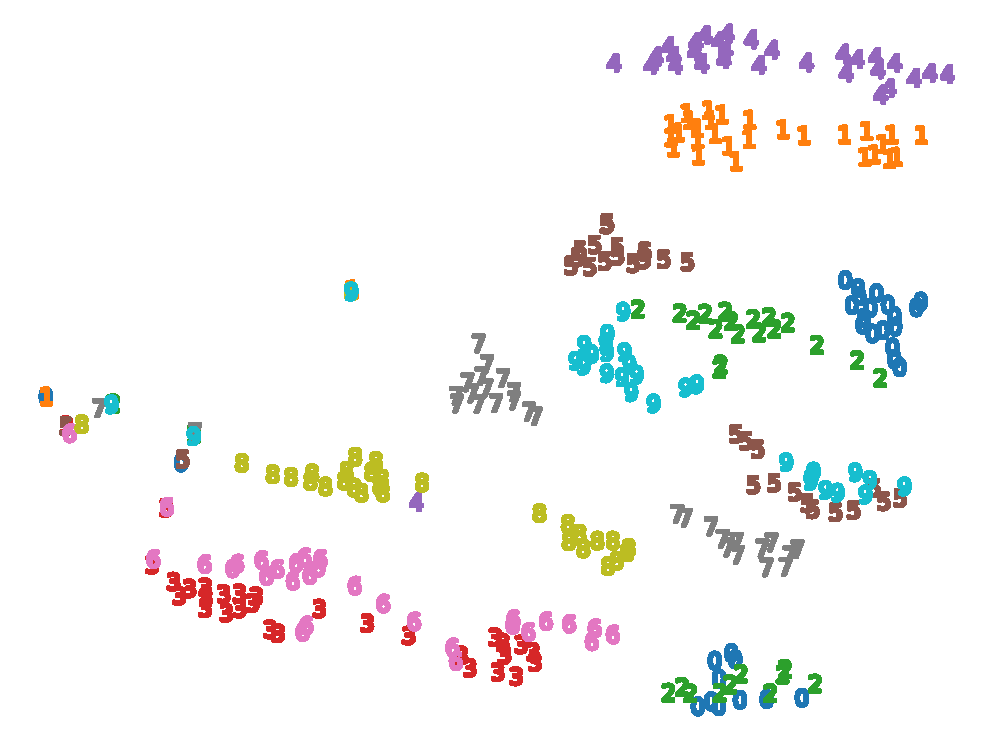} 
		\centerline{Epoch-12}
	\end{minipage}
	\hspace{10pt}
	\begin{minipage}[t]{0.14\textwidth}
		\center
		\includegraphics[width=1\textwidth]{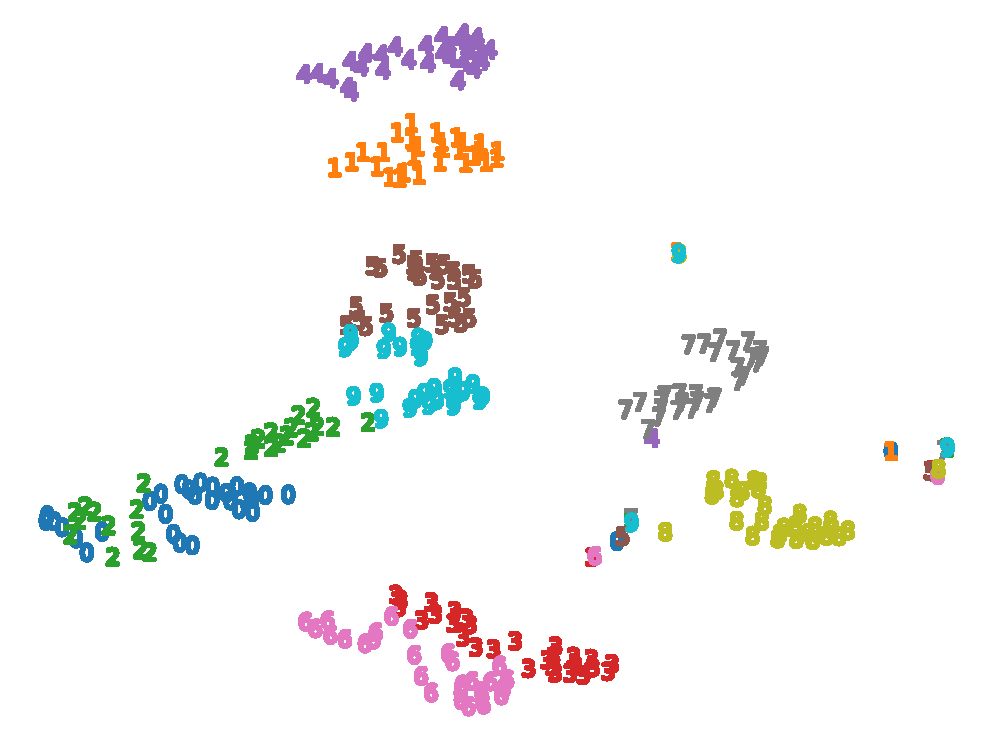} 
		\centerline{Epoch-18}
	\end{minipage}
	\hspace{10pt}
	\begin{minipage}[t]{0.14\textwidth}
		\center
		\includegraphics[width=1\textwidth]{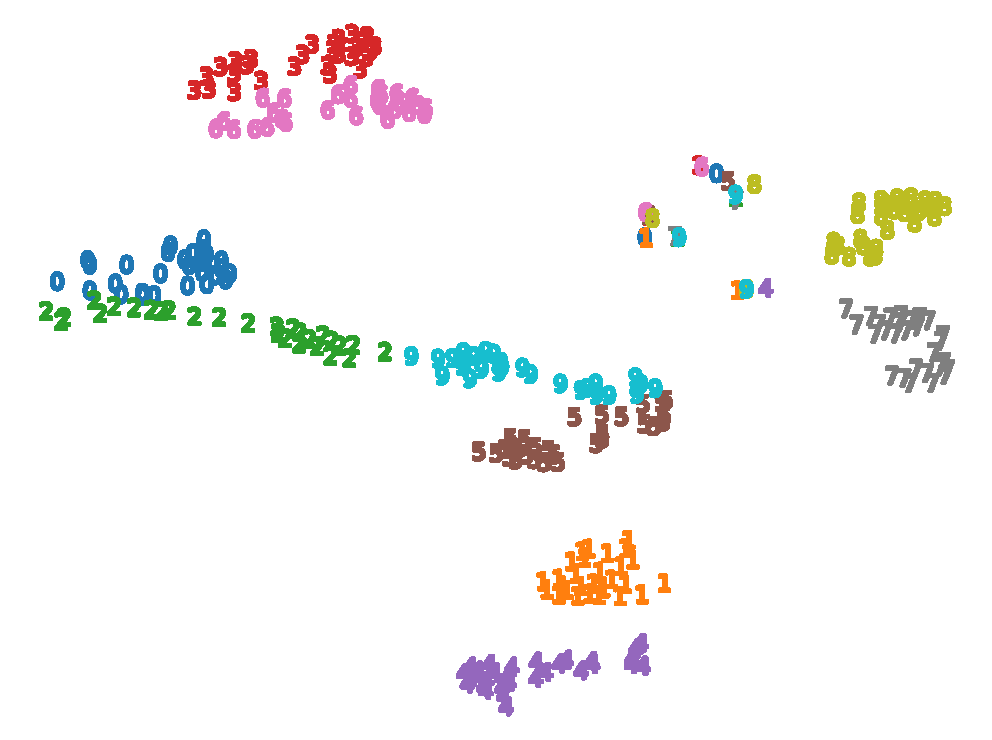} 
		\centerline{Epoch-24}
	\end{minipage}
	\hspace{10pt}
	\begin{minipage}[t]{0.14\textwidth}
		\center
		\includegraphics[width=1\textwidth]{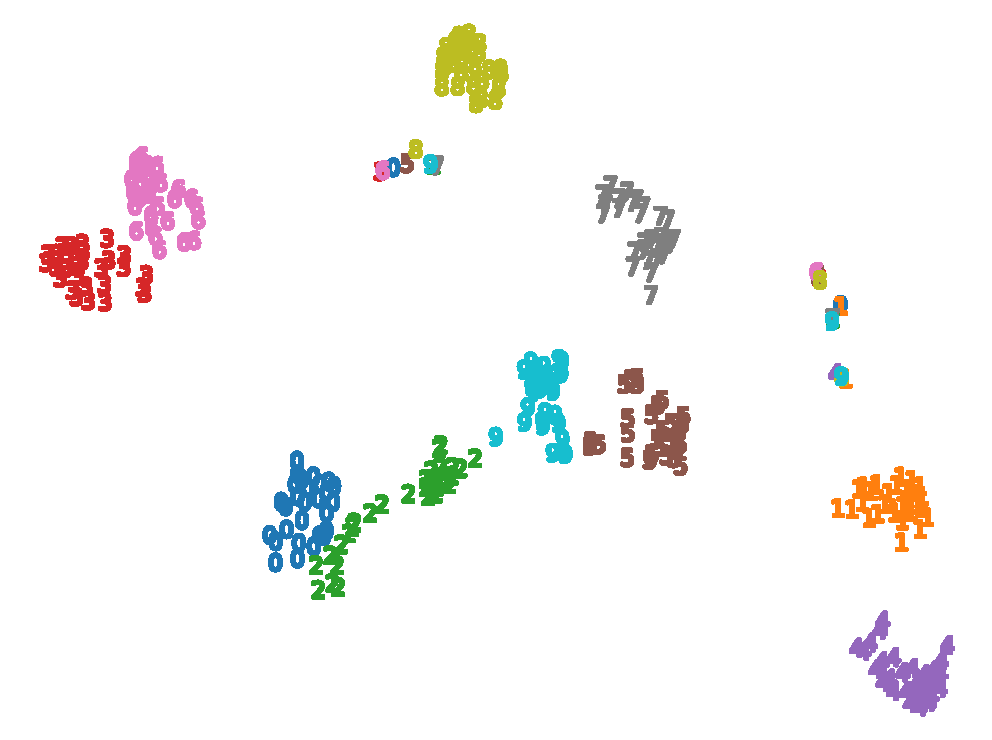} 
		\centerline{Final}
	\end{minipage}
	\caption{Visualization of partial embeddings learned by DCGL on YaleB.}
	\label{fig:tsneYaleB}
\end{figure*}

\subsection{Overall Model and Optimization}

The overall loss function of DCGL is
\begin{equation}
	\label{eq:overall_loss}
	\begin{array}{c}
		{\cal L} = {{\cal L}_{AE}} + {{\cal L}_{FL}} + \alpha {{\cal L}_{GL}} + \beta {{\cal L}_{CL}},
	\end{array}
\end{equation}
where $\alpha$ and $\beta$ are the trade-off parameters.

The optimizer for DCGL is Adam \cite{Adam}, and the corresponding parameters are predefined with the authors' recommendation. The temperature in contrastive learning is set as 0.5. The parameter $\lambda$ in Eq. (\ref{eq:gdg_ppr}) is fixed to 0.2. The convergence condition of DCGL is that the neighbor amount of LPG reaches the upper bound. Algorithm \ref{alg:dcgl} displays the complete flow of DCGL, and the final result is obtained by performing NCut \cite{NCut} on the converged LPG.

\section{Experiments}

\subsection{Benchmark Datasets}

Seven publicly available datasets are collected as benchmarks, including regular record types TOX-171 and Isolet, image types ORL, YaleB, PIE and USPS, and text type TR41. In YaleB, only the previous 30 images of each class are retained. All samples are normalized with $\ell_{2}$ norm. The details are exhibited in Table \ref{tab:datasets}.

\subsection{Evaluation Metrics}

Two metrics named accuracy (ACC) and normalized mutual information (NMI) are used to evaluate the clustering performance. The mathematical formulas of ACC and NMI can be found in \cite{l21NMF}. Higher metric values indicate better clustering performance.

\subsection{Performance Comparison}

\subsubsection{Competitors.} Thirteen state-of-the-art clustering methods are applied for performance comparison, including six traditional algorithms $k$-means \cite{kmeans}, STSC \cite{SSC_sc}, CAN \cite{CAN}, PCAN \cite{CAN}, RGC \cite{RGC} and GEMMF \cite{EMMF}, and seven deep models DAEGC \cite{DAEGC}, DCRN \cite{DCRN}, DEC \cite{DEC}, VGAE \cite{VGAE}, SpectralNet \cite{SpectralNet}, SUBLIME \cite{SUBLIME} and AdaGAE \cite{AdaGAE}. 

Except for DAEGC and DCRN, the other methods have been attested to be appropriate for non-graph-type data clustering. All codes are downloaded from online resources uploaded by authors. 

\subsubsection{Settings.}  The hyper-parameters of competitors are set as the recommendations of the authors. For example, the graph regularization weight of GEMMF is searched from $\{10^{1}, 10^{3}, 10^{5}, 10^{7}, 10^{9}\}$. The probability of dropout layers is from $\{0.1, 0.5, 0.9\}$. For those methods that require an initial graph, a KNN graph via Gaussian kernel weighting \cite{GNMF} is constructed as input, and the number of neighbors is selected from  $\{5, 6, 7, 8, 9, 10\}$. 

For DCGL, the hyper-parameters $\alpha$, $\beta$, and $\gamma$ are fixed to $1$, $10^{3}$, and $2 \times 10^{3}$ respectively. The neighbor number for LPG rises every 6 epochs. The maximum epoch number is $30$. To ensure objectivity, the random seed is fixed before code execution, and each algorithm is repeated 10 times.

\begin{table}[t]
	\centering
	\small
	\begin{tabular}{ccccc}
		\toprule
		\textbf{Dataset} & \textbf{Type} & \textbf{Samples} & \textbf{Dimension} & \textbf{Classes} \\ 
		\midrule
		\textbf{TOX-171} & Record & 171 & 5748 & 4 \\
		\textbf{ORL}	& Image & 400 & 1024 & 40 \\
		\textbf{TR41} 	& Text & 878 & 7454 & 10 \\
		\textbf{YaleB}	& Image & 1140 & 1024 & 38 \\
		\textbf{Isolet} & Record & 1560 & 617 & 26 \\
		\textbf{PIE}	& Image & 2856 & 1024 & 64 \\
		\textbf{USPS}	& Image & 9298 & 256 & 10 \\
		\bottomrule
	\end{tabular}
	\caption{Details of each benchmark dataset.}
	\label{tab:datasets}
\end{table}

\begin{table*}[t]
	\tabcolsep=0.2cm
	\centering
	\small
	\begin{tabular}{ccc|cc|cc|cc|cc|cc|cc}
		\toprule
		\makecell[c]{\multirow{2.5}{*}{Method}} & \multicolumn{2}{c}{TOX-171} & \multicolumn{2}{c}{ORL} & \multicolumn{2}{c}{TR41} & \multicolumn{2}{c}{YaleB} & \multicolumn{2}{c}{Isolet} & \multicolumn{2}{c}{PIE} & \multicolumn{2}{c}{USPS}  \\
		\cmidrule{2-15} 
		\makecell[c]{\multirow{2.5}{*}{}} & \makecell[c]{ACC} & \makecell[c]{NMI} & \makecell[c]{ACC} & \makecell[c]{NMI} & \makecell[c]{ACC} & \makecell[c]{NMI} & \makecell[c]{ACC} & \makecell[c]{NMI} & \makecell[c]{ACC} & \makecell[c]{NMI} & \makecell[c]{ACC} & \makecell[c]{NMI} & \makecell[c]{ACC} & \makecell[c]{NMI} \\
		\midrule
		\makecell[c]{$k$-means} & \makecell[c]{41.51} & \makecell[c]{12.40} & \makecell[c]{51.50} & \makecell[c]{71.07} & \makecell[c]{66.74} & \makecell[c]{67.87} & \makecell[c]{11.58} & \makecell[c]{21.48} & \makecell[c]{59.23} & \makecell[c]{75.43} & \makecell[c]{23.28} & \makecell[c]{53.56} & \makecell[c]{67.49} & \makecell[c]{61.35} \\
		\makecell[c]{STSC} & \makecell[c]{49.71} & \makecell[c]{26.35} & \makecell[c]{65.50} & \makecell[c]{77.70} & \makecell[c]{72.67} & \makecell[c]{\underline{68.16}} & \makecell[c]{49.04} & \makecell[c]{62.29}  & \makecell[c]{58.91} & \makecell[c]{76.62} & \makecell[c]{86.80} & \makecell[c]{93.51} & \makecell[c]{66.44} & \makecell[c]{\underline{80.03}} \\
		\makecell[c]{CAN} & \makecell[c]{32.75} & \makecell[c]{10.28} & \makecell[c]{56.00} & \makecell[c]{72.04} & \makecell[c]{51.59} & \makecell[c]{35.61} & \makecell[c]{39.74} & \makecell[c]{52.50} & \makecell[c]{60.77} & \makecell[c]{77.38} & \makecell[c]{91.81} & \makecell[c]{97.02} & \makecell[c]{74.10} & \makecell[c]{76.89} \\
		\makecell[c]{PCAN} & \makecell[c]{45.03} & \makecell[c]{21.66} & \makecell[c]{61.00} & \makecell[c]{74.18} & \makecell[c]{62.30} & \makecell[c]{53.97} & \makecell[c]{45.09} & \makecell[c]{49.73} & \makecell[c]{57.24} & \makecell[c]{69.87} & \makecell[c]{85.68} & \makecell[c]{87.88} & \makecell[c]{65.41} & \makecell[c]{76.16} \\
		\makecell[c]{RGC} & \makecell[c]{50.88} & \makecell[c]{28.60} & \makecell[c]{59.50} & \makecell[c]{77.11} & \makecell[c]{64.24} & \makecell[c]{64.06} & \makecell[c]{31.32} & \makecell[c]{43.68} & \makecell[c]{63.53} & \makecell[c]{77.92} & \makecell[c]{40.62} & \makecell[c]{68.29} & \makecell[c]{72.00} & \makecell[c]{78.93} \\
		\makecell[c]{GEMMF} & \makecell[c]{43.86} & \makecell[c]{13.19} & \makecell[c]{64.75} & \makecell[c]{79.29} & \makecell[c]{58.89} & \makecell[c]{57.74} & \makecell[c]{31.40} & \makecell[c]{51.84} & \makecell[c]{63.53} & \makecell[c]{78.23} & \makecell[c]{41.81} & \makecell[c]{71.35} & \makecell[c]{69.96} & \makecell[c]{66.43} \\
		\midrule
		\makecell[c]{DAEGC} & \makecell[c]{46.20} & \makecell[c]{16.90} & \makecell[c]{59.50} & \makecell[c]{76.09} & \makecell[c]{61.62} & \makecell[c]{57.87} & \makecell[c]{23.86} & \makecell[c]{42.86} & \makecell[c]{59.68} & \makecell[c]{73.11} & \makecell[c]{39.57} & \makecell[c]{71.48} & \makecell[c]{70.43} & \makecell[c]{59.77} \\
		\makecell[c]{DCRN} &  \makecell[c]{36.84} & \makecell[c]{12.89} & \makecell[c]{34.25} & \makecell[c]{60.10} & \makecell[c]{62.30} & \makecell[c]{54.14} & \makecell[c]{15.61} & \makecell[c]{34.03} & \makecell[c]{15.06} & \makecell[c]{36.99} & \makecell[c]{17.75} & \makecell[c]{47.59} & \makecell[c]{18.12} & \makecell[c]{5.27} \\
		\midrule
		\makecell[c]{DEC} & \makecell[c]{49.71} & \makecell[c]{23.65} & \makecell[c]{49.25} & \makecell[c]{69.33} & \makecell[c]{23.46} & \makecell[c]{10.88} & \makecell[c]{21.84} & \makecell[c]{35.60} & \makecell[c]{52.76} & \makecell[c]{62.85} & \makecell[c]{30.22} & \makecell[c]{59.62} & \makecell[c]{76.34} & \makecell[c]{66.72} \\
		\makecell[c]{VGAE}  & \makecell[c]{49.71} & \makecell[c]{22.99} & \makecell[c]{56.75} & \makecell[c]{74.99} & \makecell[c]{\underline{72.78}} & \makecell[c]{64.25} & \makecell[c]{27.19} & \makecell[c]{40.60} & \makecell[c]{63.59} & \makecell[c]{77.06} & \makecell[c]{38.41} & \makecell[c]{63.17} & \makecell[c]{66.66} & \makecell[c]{60.78} \\
		\makecell[c]{SpectralNet} & \makecell[c]{46.78} & \makecell[c]{20.55} & \makecell[c]{53.50} & \makecell[c]{72.66} & \makecell[c]{34.62} & \makecell[c]{21.63} & \makecell[c]{19.91} & \makecell[c]{33.66} & \makecell[c]{56.03} & \makecell[c]{66.21} & \makecell[c]{79.03} & \makecell[c]{87.87} & \makecell[c]{49.31} & \makecell[c]{39.70} \\
		\makecell[c]{SUBLIME} & \makecell[c]{\underline{51.46}} & \makecell[c]{26.87} & \makecell[c]{68.75} & \makecell[c]{82.29} & \makecell[c]{65.49} & \makecell[c]{66.85} & \makecell[c]{47.72} & \makecell[c]{62.41} & \makecell[c]{64.41} & \makecell[c]{76.64} & \makecell[c]{82.35} & \makecell[c]{90.15} & \makecell[c]{59.14} & \makecell[c]{64.50} \\
		\makecell[c]{AdaGAE} & \makecell[c]{49.71} & \makecell[c]{\underline{30.60}} & \makecell[c]{\underline{69.50}} & \makecell[c]{\underline{84.40}} & \makecell[c]{55.81} & \makecell[c]{59.46} & \makecell[c]{\underline{67.02}} & \makecell[c]{\underline{80.68}} & \makecell[c]{\underline{69.49}} & \makecell[c]{\underline{80.83}} & \makecell[c]{\underline{97.97}} & \makecell[c]{\underline{98.18}} & \makecell[c]{\underline{76.44}} & \makecell[c]{74.63} \\
		\midrule
		\makecell[c]{DCGL} & \makecell[c]{\textbf{51.46}} & \makecell[c]{\textbf{34.75}} & \makecell[c]{\textbf{75.50}} & \makecell[c]{\textbf{86.91}} & \makecell[c]{\textbf{73.23}} & \makecell[c]{\textbf{69.80}} & \makecell[c]{\textbf{81.93}} & \makecell[c]{\textbf{86.18}} & \makecell[c]{\textbf{71.73}} & \makecell[c]{\textbf{80.85}} & \makecell[c]{\textbf{99.86}} &  \makecell[c]{\textbf{99.87}} & \makecell[c]{\textbf{78.32}} & \makecell[c]{\textbf{80.87}} \\
		\bottomrule
	\end{tabular}
	\caption{Clustering performance of different methods on benchmark datasets. Bold and underlined values indicate the optimal and sub-optimal results respectively.}
	\label{tab:comparsion_acc}
\end{table*}

\subsubsection{Results.} The best clustering results of each algorithm are recorded in Table \ref{tab:comparsion_acc}. We summarize the viewpoints that 1) STSC, GEMMF, VGAE, and SpectralNet perform worse than other graph-based methods relatively. They keep the graph structure fixed during iteration, so the final performance is limited by the initial graph quality. This reveals that adaptive graph learning is beneficial to clustering; 2) both DAEGC and DCRN provide unsatisfactory results. These models assume that the initial graph is highly reliable, and learn the cluster embeddings mainly based on the adjacency relation. The experimental results attest that such graph-type data clustering models are not suitable for general clustering tasks; 3) the deep models for general data clustering, i.e., SUBLIME, AdaGAE, and DCGL, realize a noticeable improvement. This verifies the potential of deep graph-based clustering. The good performance of SUBLIME and DCGL also demonstrates the feasibility of contrastive learning on processing non-vision data; 4) DCGL achieves better clustering compared to GCN-based competitors, which reflects the effects of restraining representation collapse and mining the original features. Besides, the difference between DCGL and GAE-based models mainly dates from the siamese graph convolution module that bootstraps a clear cluster structure gradually. With the cooperation of clustering-oriented mechanisms, DCGL is guided to preserve clustering-relevant information, and presents superior performance compared to other algorithms.

\subsection{Visualization Analysis}

To display the effect of DCGL intuitively, we use $t$-SNE \cite{tSNE} to visualize the embeddings from $\mathrm{GCN}_1$ when clustering YaleB. Only the front 10 clusters are processed for ease of observation. The visualization result is shown in Fig. \ref{fig:tsneYaleB}, where each point is plotted as its actual label value.  It can be seen that the samples of the same category are gradually compact with iteration, which indicates the practicability of DCGL on data clustering.

\subsection{Ablation Studies}

In this part, we first validate the effectiveness of two contrastive learning modules in DCGL, and then explore the impact of clustering-oriented mechanisms. Afterward, we analyze the impact of hyper-parameters.

\subsubsection{Impact of Contrastive Learning.}

\begin{figure}[t]
	\center
	\begin{minipage}{0.21\textwidth}
		\center
		\includegraphics[width=1\textwidth]{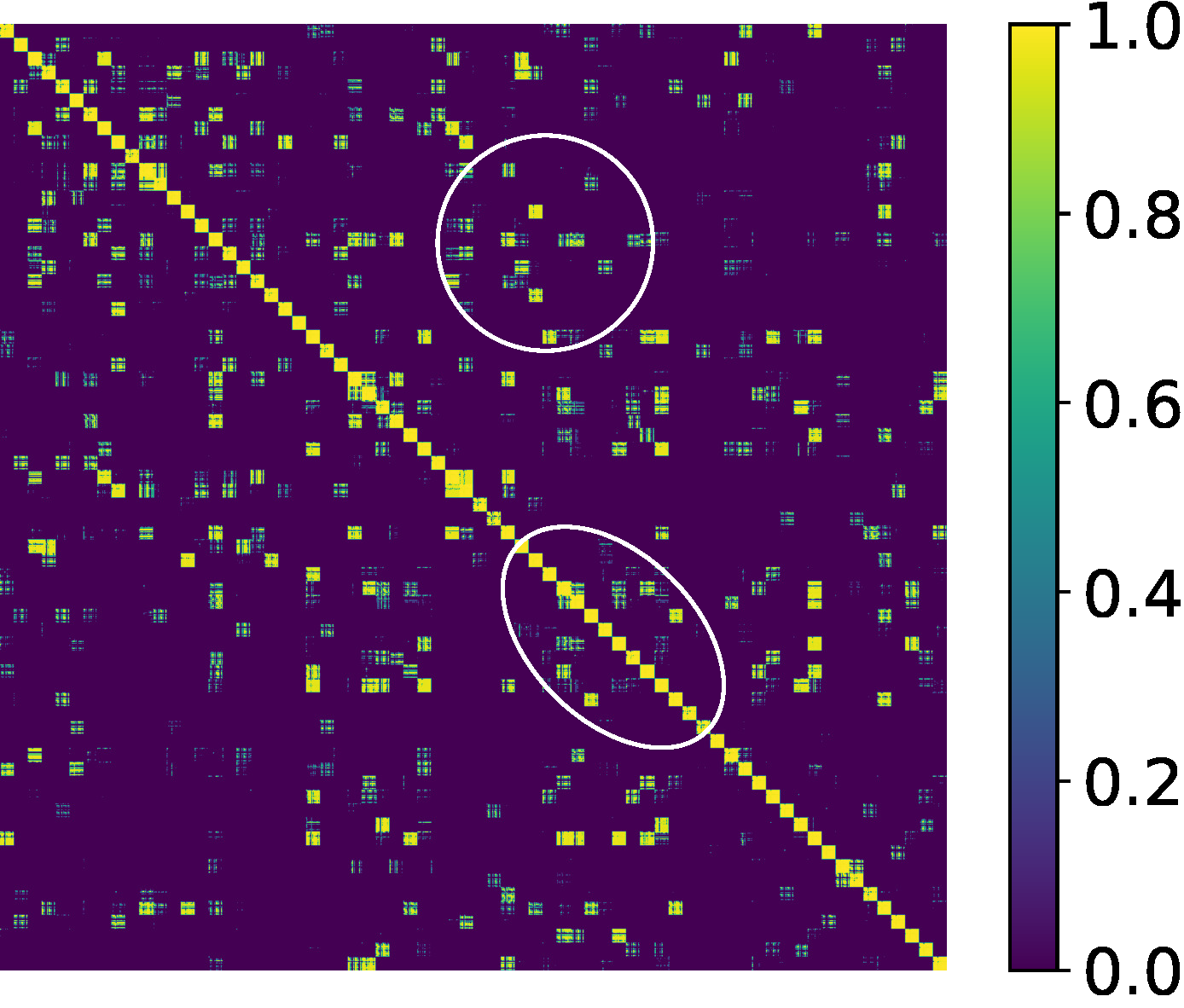} 
		\centerline{DCGL-w/F}
	\end{minipage}
	\hspace{10pt}
	\begin{minipage}{0.21\textwidth}
		\center
		\includegraphics[width=1\textwidth]{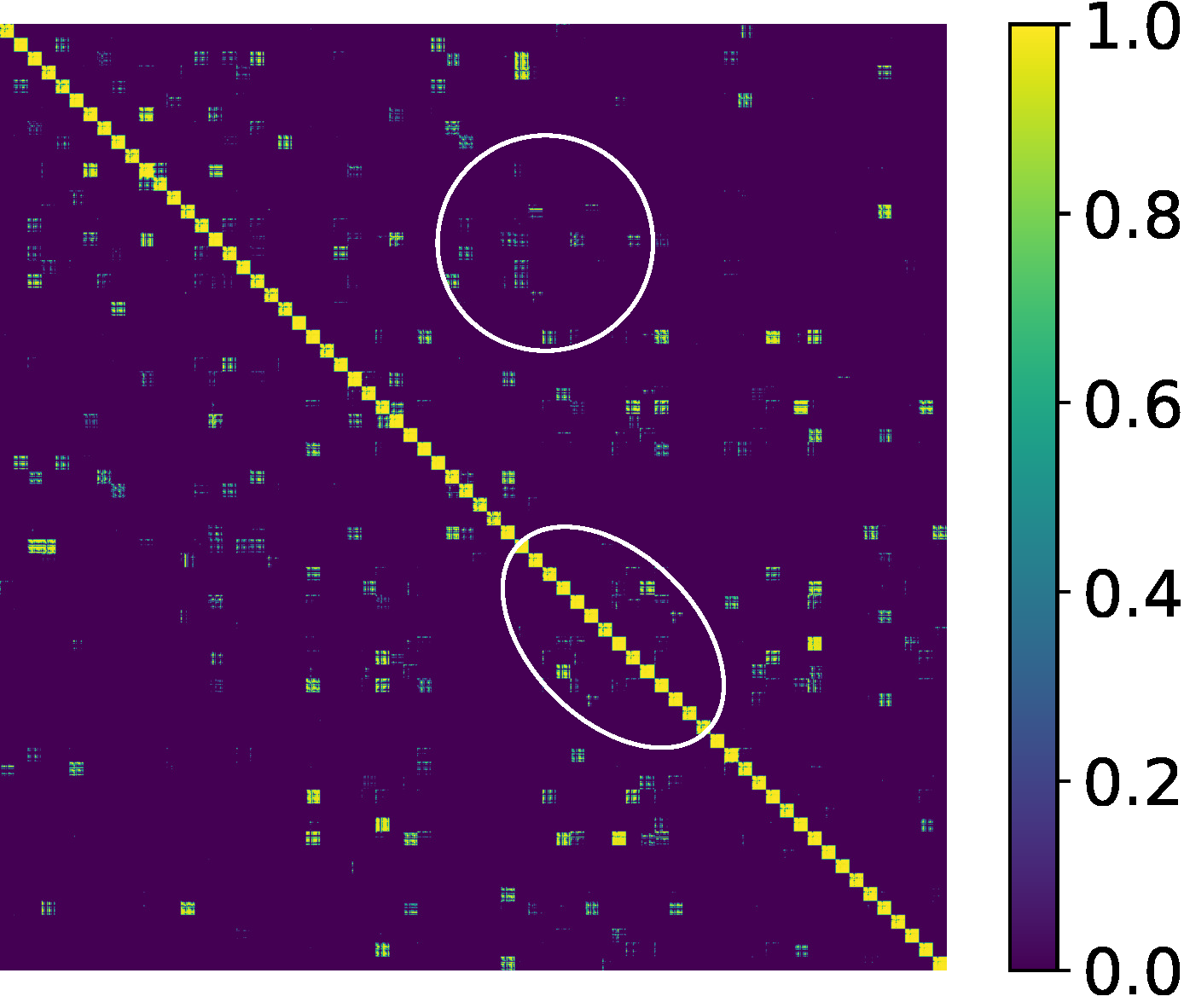} 
		\centerline{DCGL}
	\end{minipage}
	\caption{Ablation results of feature-level contrastive learning on PIE.}
	\label{fig:ablation_fl}
\end{figure}

Define DCGL-w/F and DCGL-w/C as the DCGL variants without feature-level and cluster-level contrastive learning respectively. Since the outputs of $\mathrm{GCN}_2$ and $\mathrm{GCN}_3$ are only related to cluster-level contrastive learning, DCGL-w/C means the entire expurgation of the siamese graph convolution module. Table \ref{tab:ablation_com} shows the ablation comparisons. Obviously, DCGL still presents the best clustering performance, which proves the effectiveness of each contrastive learning strategy. 

The performance of DCGL-w/F is deteriorated compared to DCGL, which demonstrates the discriminability of the embeddings learned by DCGL. For a better illustration, we calculate the cosine similarity matrix of the data representation from $\mathrm{GCN}_1$. The heatmap on Pie is displayed in Fig. \ref{fig:ablation_fl}, where only the large values are preserved to facilitate observation. In the result of DCGL, the nodes across diverse classes have lower similarity, so we can conclude that feature-level contrastive learning is beneficial to improving discriminability.

\begin{figure}[t]
	\center
	\begin{minipage}[t]{0.21\textwidth}
		\center
		\includegraphics[width=1\textwidth]{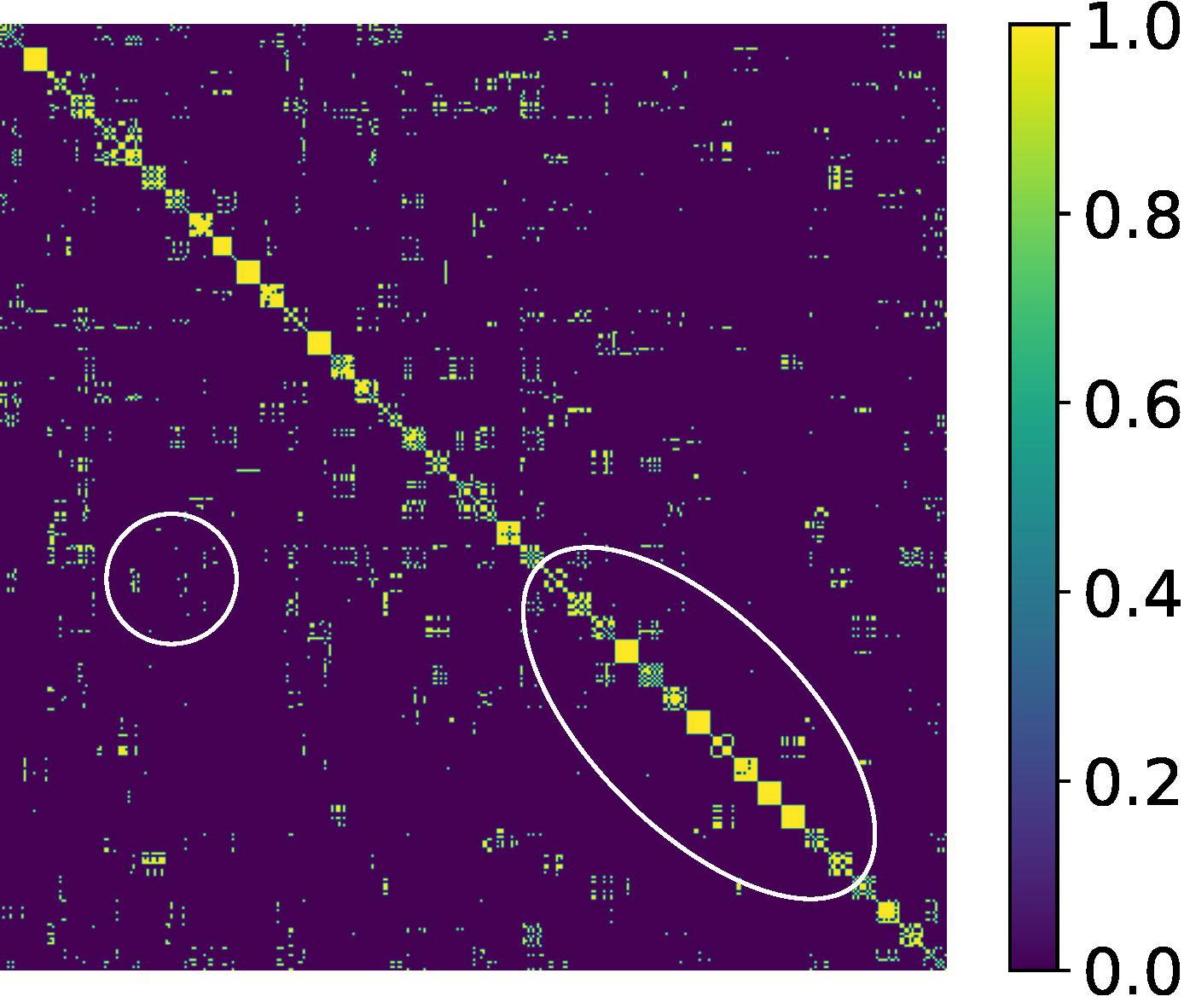} 
		\centerline{DCGL-w/C}
	\end{minipage}
	\hspace{10pt}
	\begin{minipage}[t]{0.21\textwidth}
		\center
		\includegraphics[width=1\textwidth]{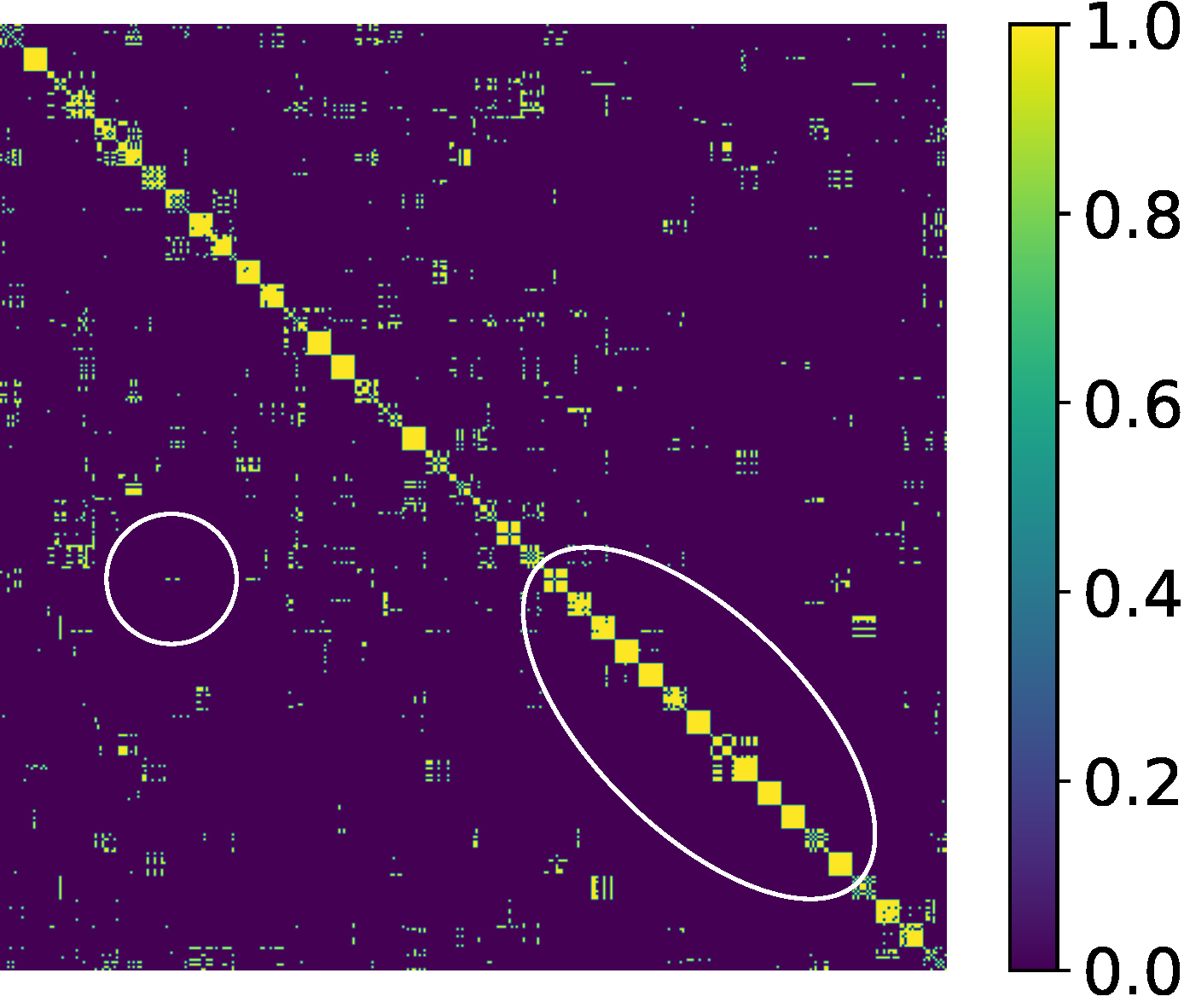} 
		\centerline{DCGL}
	\end{minipage}
	\caption{Ablation results of cluster-level contrastive learning on ORL.}
	\label{fig:ablation_cl}
\end{figure}

DCGL exceeds DCGL-w/C on all benchmark datasets, which means the graph generated by DCGL maintains a clearer cluster structure. To verify the efficiency of cluster-level contrastive learning on graph optimization, we plot the converged LPGs from DCGL-w/C and DCGL on ORL in Fig. \ref{fig:ablation_cl}, where all non-zero elements are substituted with 1 to emphasize the adjacency relation. It can be seen that the graph obtained by DCGL exhibits a more distinct block diagonal structure, indicating more links between samples of the same category. 

\begin{table}[t]
	\centering
	\small
	\begin{tabular}{c|c|ccc}
		\toprule
		Dataset & Metric & DCGL-w/F & DCGL-w/C & DCGL \\ 
		\midrule
		\multirow{2}{*}{TOX-171} & ACC & 47.95 & 42.11 & \textbf{51.46} \\ 
		\multirow{2}{*}{} & NMI & 24.17 & 24.84 & \textbf{34.75} \\ 
		\midrule
		\multirow{2}{*}{ORL} & ACC & 70.50 & 64.00 & \textbf{75.50} \\ 
		\multirow{2}{*}{} & NMI & 83.72 & 81.17 & \textbf{86.91} \\ 
		\midrule
		\multirow{2}{*}{TR41} & ACC & 61.73 & 63.21 & \textbf{73.23} \\ 
		\multirow{2}{*}{} & NMI & 64.85 & 65.39 & \textbf{69.80} \\ 
		\midrule
		\multirow{2}{*}{YaleB} & ACC & 75.18 & 77.54 & \textbf{81.93} \\ 
		\multirow{2}{*}{} & NMI & 82.89 & 84.23 & \textbf{86.18} \\ 
		\midrule
		\multirow{2}{*}{Isolet} & ACC & 69.23 & 67.63 & \textbf{71.73} \\ 
		\multirow{2}{*}{} & NMI & 79.63 & 78.92 & \textbf{80.85} \\ 
		\midrule
		\multirow{2}{*}{PIE} & ACC & 99.68 & 98.35 & \textbf{99.86} \\ 
		\multirow{2}{*}{} & NMI & 99.69 & 99.65 & \textbf{99.87} \\ 
		\midrule
		\multirow{2}{*}{USPS} & ACC & 76.75 & 76.98 & \textbf{78.32} \\ 
		\multirow{2}{*}{} & NMI & 80.19 & 79.09 & \textbf{80.87} \\ 
		\bottomrule
	\end{tabular}
	\caption{Ablation results of each contrastive learning module in DCGL. Bold values indicate the optimal results.}
	\label{tab:ablation_com}
\end{table}

\begin{figure}[t]
	\center
	\begin{minipage}{0.21\textwidth}
		\center
		\includegraphics[width=1\textwidth]{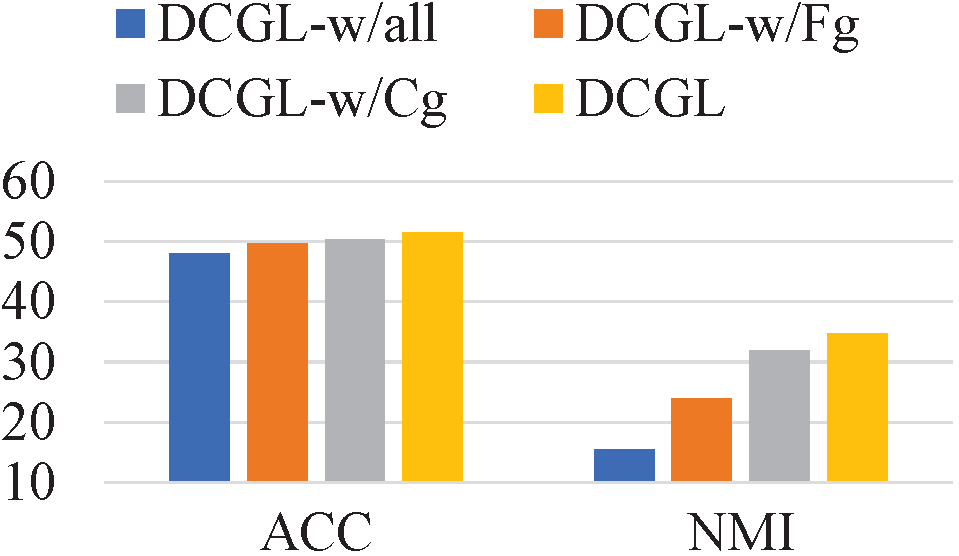} 
		\centerline{TOX-171}
	\end{minipage}
	\hspace{10pt}
	\begin{minipage}{0.21\textwidth}
		\center
		\includegraphics[width=1\textwidth]{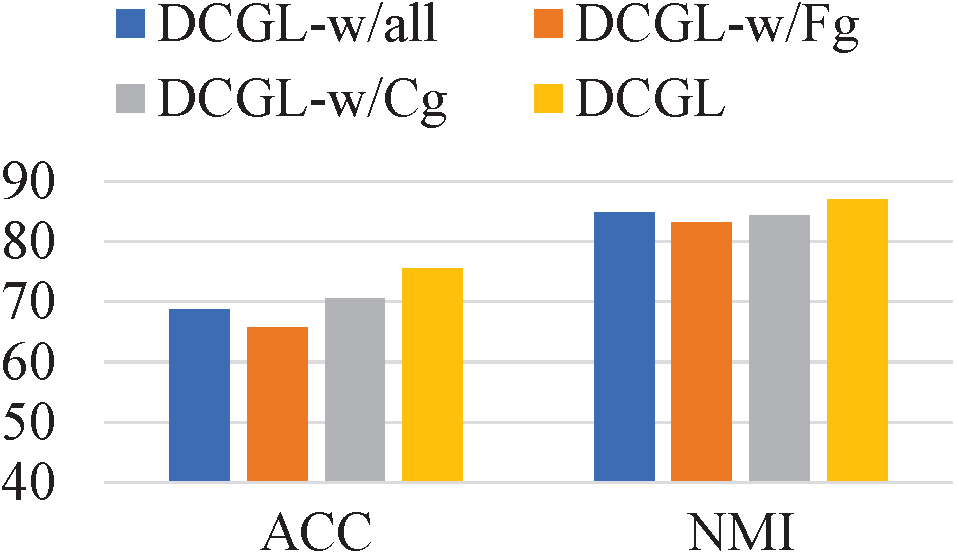} 
		\centerline{ORL}
	\end{minipage}
	\center
	\begin{minipage}{0.21\textwidth}
		\center
		\includegraphics[width=1\textwidth]{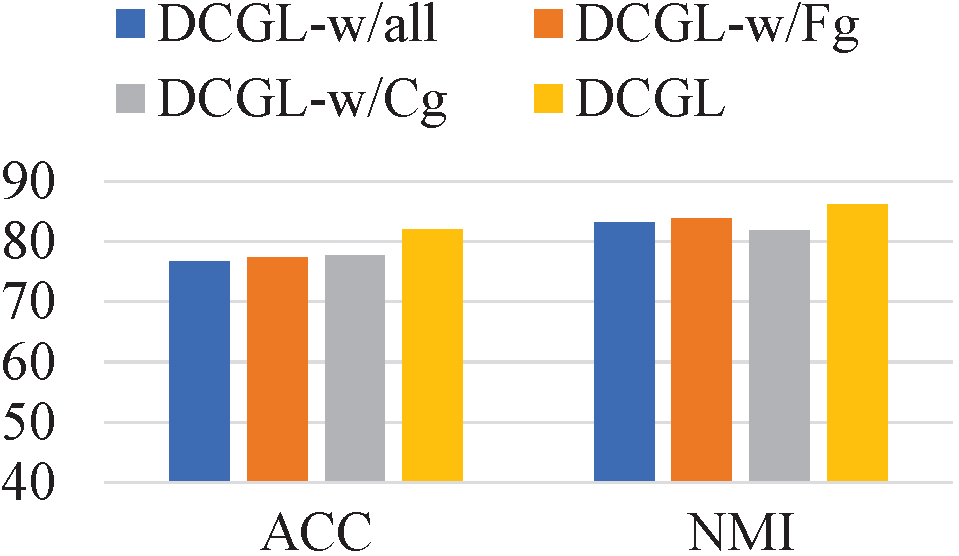} 
		\centerline{YaleB}
	\end{minipage}
	\hspace{10pt}
	\begin{minipage}{0.21\textwidth}
		\center
		\includegraphics[width=1\textwidth]{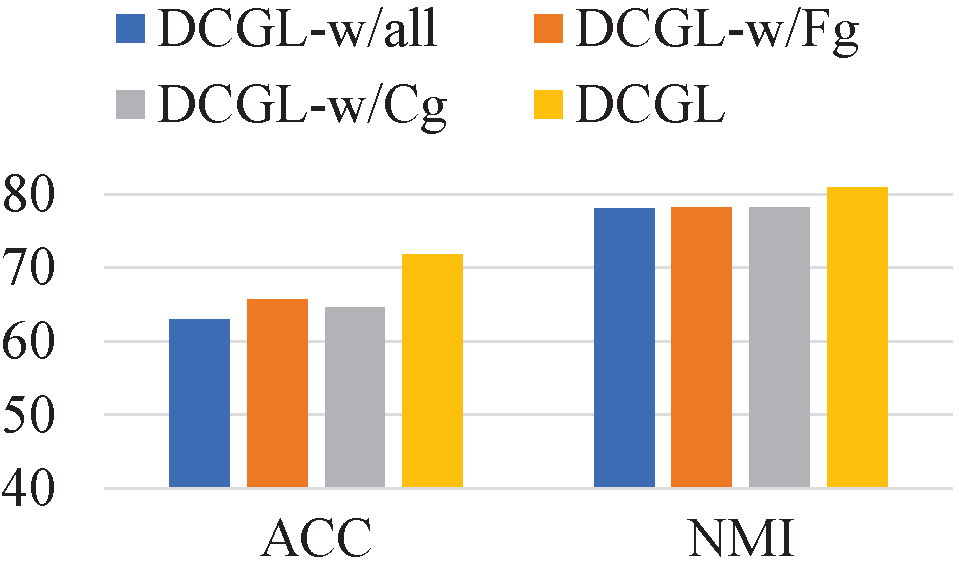} 
		\centerline{Isolet}
	\end{minipage}
	\caption{Ablation results of clustering guidance on four datasets.}
	\label{fig:ablation_gd}
\end{figure}

\subsubsection{Impact of Clustering Guidance.} In the proposed contrastive learning strategies, feature-level and cluster-level guidance are introduced into the construction of sample pairs and the selection of anchors. To investigate the effectiveness of the above clustering guidance, we design three variants of DCGL with different contrastive learning approaches, termed as DCGL-w/Fg, DCGL-w/Cg, and DCGL-w/all. Among them, DCGL-w/Fg discards the feature-level guidance, and regards all non-corresponding nodes of the anchor in the other branch as negative samples in feature-level contrastive learning. DCGL-w/Cg removes the cluster-level guidance by taking the output embeddings of $\mathrm{GCN}_2$ and $\mathrm{GCN}_3$ as anchors directly. DCGL-w/all has the above two alterations.

Fig. \ref{fig:ablation_gd} presents the comparison results. We can deduce that 1) the evident degradation of DCGL-w/all confirms the significance of clustering guidance; 2) the decline of DCGL-w/Fg implies that the discriminability is decreased without the feature-level guidance; 3) the disparity between DCGL-w/Cg and DCGL demonstrates that contrastive learning within the cluster space is helpful to perceive a clear cluster distribution.

\subsubsection{Sensitivity of Hyper-Parameters $\alpha$ and $\beta$.} 
\begin{figure}[t]
	\center
	\begin{minipage}[t]{0.22\textwidth}
		\centering
		\includegraphics[width=1\textwidth]{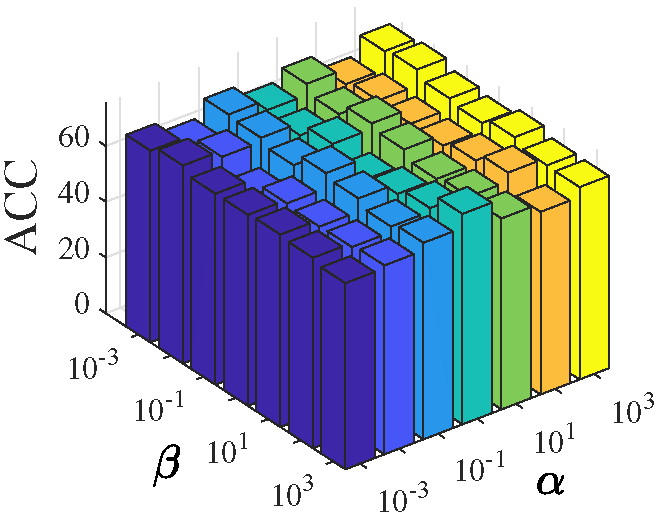}
		\centerline{ORL}
	\end{minipage}
	\hspace{5pt}
	\begin{minipage}[t]{0.22\textwidth}
		\centering
		\includegraphics[width=1\textwidth]{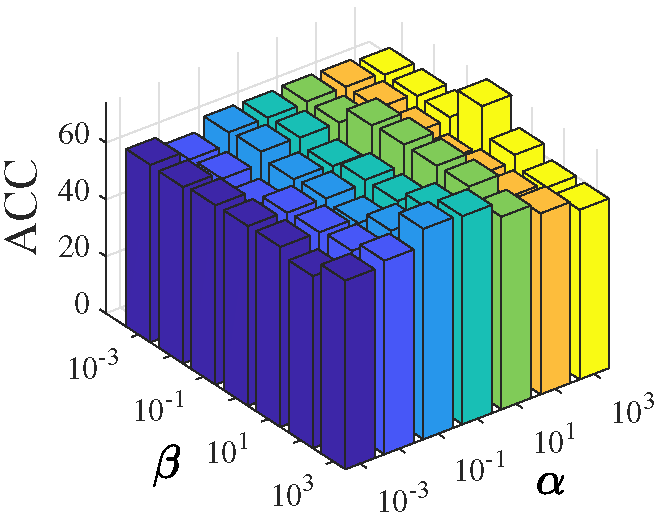}
		\centerline{TR41}
	\end{minipage}
	\center
	\begin{minipage}[t]{0.22\textwidth}
		\centering
		\includegraphics[width=1\textwidth]{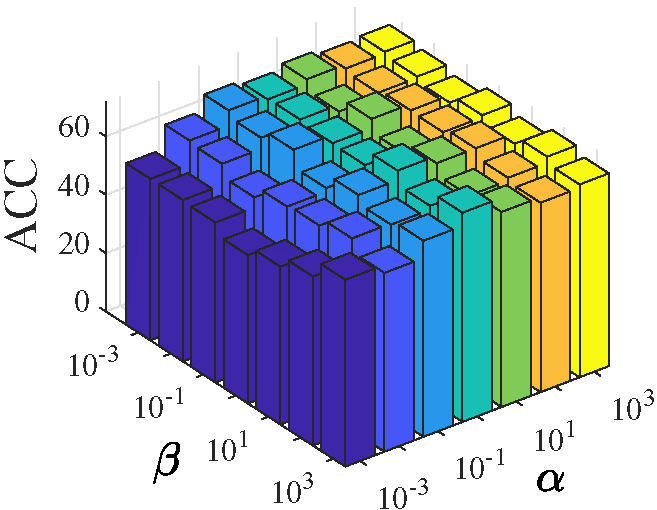}
		\centerline{Isolet}
	\end{minipage}
	\hspace{5pt}
	\begin{minipage}[t]{0.22\textwidth}
		\centering
		\includegraphics[width=1\textwidth]{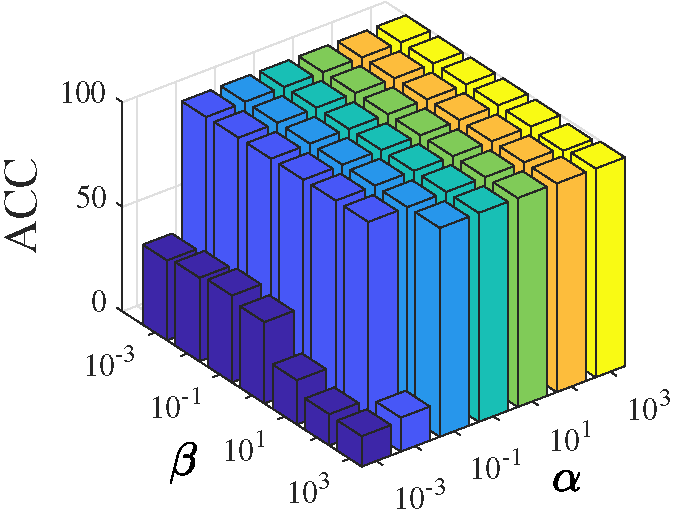}
		\centerline{PIE}
	\end{minipage}
	\caption{ACC of DCGL with different hyper-parameters $\alpha$ and $\beta$.}
	\label{fig:par_sensitivity}
\end{figure}

Finally, we focus on the influence of trade-off parameters $\alpha$ and $\beta$. We tune both $\alpha$ and $\beta$ over the range of $\{10^{-3}, 10^{-2}, 10^{-1}, 1, 10^{1}, 10^{2}, 10^{3}\}$, and keep other experimental environments. Fig. \ref{fig:par_sensitivity} exhibits the clustering ACC of DCGL with different parameter combinations. It is observed that 1) the performance under too small $\alpha$ and $\beta$ is not outstanding, which manifests the failure of related mechanisms, and again attests to the effectiveness of the designed modules; 2) the influence of $\alpha$ is more salient relatively, because $\alpha$ is associated with graph learning and the final cluster partitioning directly; 3) the setting of moderate $\alpha$ and larger $\beta$ is likely to bring satisfactory clustering results. Overall, DCGL is not particularly sensitive to parameter selection in an appropriate range.

\section{Conclusion}

In this paper, we establish the Deep Contrastive Graph Learning (DCGL) model for non-graph data clustering. DCGL adopts a pseudo-siamese network to learn both the structural and attributed representations. Two clustering-oriented contrastive learning strategies are also designed to capture clustering-relevant information. Concretely, feature-level contrastive learning is performed to push each node away from the centroids of other clusters, to retain the discriminative features of each class. Besides, cluster-level contrastive learning is conducted to search an explicit centroid distribution shared by both the learned local and global graphs, such that a clear cluster structure is achieved. Extensive experiments on benchmark datasets demonstrate the superiority of DCGL against state-of-the-art competitors.

\section{Acknowledgments}
This work was supported by the National Key Research and Development Program of China (No. 2022YFC2808000) and by the National Natural Science Foundation of China (No. 61871470).

\bibliography{aaai24}

\end{document}